\documentclass{article}

\PassOptionsToPackage{numbers}{natbib}

\usepackage[preprint]{neurips_2026}

\usepackage[utf8]{inputenc}
\usepackage[T1]{fontenc}
\usepackage{hyperref}
\usepackage{url}
\usepackage{booktabs}
\usepackage{amsfonts}
\usepackage{amsmath}
\usepackage{amssymb}
\usepackage{nicefrac}
\usepackage{microtype}
\usepackage{xcolor}
\usepackage{graphicx}
\usepackage{tabularx}
\usepackage{algorithm}
\usepackage{algpseudocode}
\usepackage{graphicx}
\usepackage{subcaption}
\usepackage{adjustbox}
\usepackage{xcolor}
\usepackage{float}

\newcommand{\exampletext}[1]{%
\begin{minipage}[t]{0.40\textwidth}%
\vspace{0pt}%
\scriptsize\ttfamily%
\setlength{\parindent}{0pt}%
\setlength{\parskip}{0pt}%
\linespread{0.92}\selectfont%
#1%
\end{minipage}%
}
\newcommand{\exampleimage}[1]{%
\begin{minipage}[t]{0.38\textwidth}%
\vspace*{0pt}\vspace*{0.4em}%
\centering\includegraphics[width=\linewidth]{#1}%
\end{minipage}%
}

\title{LinTree: Improving LLM Reasoning with Explicitly Structured Search Histories}

\author{
Liwei Kang$^{1}$ \quad Yee Whye Teh$^{2}$ \quad Wee Sun Lee$^{1}$\thanks{Correspondence to Wee Sun Lee, email: \texttt{dcsleews@nus.edu.sg}.} \vspace{+5pt} \\
$^{1}$National University of Singapore \ \ 
$^{2}$University of Oxford
}

\begin{document}

\maketitle

\begin{abstract}
Large language models (LLMs) often solve reasoning problems by generating intermediate traces that explore and revise partial solutions. From a search perspective, these traces can be viewed as linearized search trees, where the model extends a partial solution, abandons it when it fails, and backtracks to try alternatives. Compared with traditional heuristic-guided search, such a policy has a potential advantage: it conditions on the whole search trace rather than only on the current local state. We first test whether LLMs utilize this advantage by comparing trace-conditioned reasoning policies against best-first search equipped with an LLM heuristic that only observes the current local state. Across three controlled reasoning environments, Blocks World, grid Navigation, and Sokoban, we find that raw access to search history alone is not enough to reliably outperform heuristic search. We then study one possible reason: in LLM reasoning traces, the underlying search tree is only implicitly represented, and when the model backtracks or switches branches, the trace does not explicitly identify which earlier search state is being revisited. We show that adding simple parent pointers to explicitly represent the linearized tree (\textbf{LinTree}) structure improves both task performance and search efficiency relative to implicit reasoning models and LLM-heuristic-guided search. These results suggest that search history becomes most useful when its tree structure is made explicit, motivating more structure-aware representations for LLM reasoning.
\end{abstract}

\section{Introduction}\label{sec:intro}

Large language models (LLMs) have demonstrated remarkable reasoning capabilities, largely driven by techniques that encourage intermediate computation, such as Chain-of-Thought (CoT) prompting and self-reflection \citep{wei2022chain, nye2021show, renze2024self, madaan2023self}. Recent work further highlights that reasoning can be viewed as an implicit search process over a tree of thoughts \citep{yao2023tree, sel2023algorithm, katz2024thought}. Rather than producing an answer in a single pass, the model can generate intermediate steps, try alternative approaches, and revise earlier choices \citep{chen2024toward, shinn2023reflexion, chen2024boosting}. These developments motivate a search-based view of reasoning, in which the model is implicitly maintaining a tree structure in its chain of thoughts.

Under this search-based view, a reasoning trace can be seen as a linearized search tree, where each step extends an existing branch, revises a previous choice, or returns to an earlier point to explore an alternative \citep{yao2023tree, sel2023algorithm, chen2024toward}. Compared with traditional heuristic-guided search \citep{hart1968formal, pearl1983heuristics}, this gives the LLM a potential advantage: instead of choosing the next expansion by scoring frontier nodes using local state information, the LLM generates the next search step conditioned on the entire preceding search trace. With this richer information, the model could in principle use evidence gathered on one branch to guide choices on another \citep{steinmetz2016towards}, avoid revisiting neighborhoods that earlier expansions have already shown to be unproductive, and coordinate search using non-local context that a local heuristic never sees. Under this view, the advantage of LLM search is not simply that it can search and self-reflect, but also that it conditions on the entire search history while doing so.

This framing suggests the first question we study: does conditioning on the whole trace translate into stronger search? We test this by comparing reasoning models that observe the whole search trace against a best-first search algorithm equipped with a local-state LLM heuristic that sees only the current state and goal. Across three controlled reasoning environments, Blocks World, grid Navigation, and Sokoban, we find that trace access alone is not enough: the reasoning model does not reliably outperform the local-state heuristic baseline.

This result raises a follow-up question: why does trace access not help more? One explanation lies in how the trace is presented. Although a reasoning trace can be viewed as a linearization of a search tree \citep{besta2024graph, sel2023algorithm}, that tree structure is usually left implicit. In practice, when an LLM abandons one line of reasoning and tries another, the trace may contain a natural-language cue such as ``let me try a different approach'' \citep{guo2025deepseek, chen2024toward, jaech2024openai, li2025llms}, but it typically does not explicitly identify which earlier search state is being revisited. The history therefore contains evidence about exploration and backtracking, but its tree topology has to be inferred from context \citep{yao2023tree, chen2024toward, wang2025don}. This motivates our second question: can LLMs learn better search policies when the search-tree structure is made explicit in their training traces? We study this through a minimal change in trace representation: we compare one format in which each search step records the frontier state being expanded, thus exposing the tree topology, with another format in which it does not. Both formats are derived from the same underlying searches and trained with the same supervised fine-tuning (SFT) followed by reinforcement learning (RL) pipeline, so differences between them primarily reflect the effect of exposing tree topology rather than differences in trace access, optimization, or task difficulty. Intuitively, this turns a loosely narrated reasoning trace into a serialized search tree with clear topology, making the available history easier for the model to use.

We study these questions empirically in three controlled reasoning environments, Blocks World \citep{valmeekam2022large}, Grid Navigation \citep{mittal2025learning}, and Sokoban \citep{boxobanlevels}, where the search tree is well defined and search efficiency can be measured exactly. Together, our experiments test whether conditioning on the whole search trace is sufficient to improve search over a local-state LLM heuristic, and whether explicitly representing the search-tree structure makes that trace more useful. Our findings are summarized as follows:
\begin{itemize}
    \item While LLM reasoning models generate reasoning steps conditioned on the whole search trace, they do not fully exploit this information when the tree structure is left implicit: the reasoning policy does not reliably outperform a local-state heuristic baseline despite having access to more information.
    \item Making the tree topology of the search trace explicit changes this: trained the same way, policies that explicitly annotate the search tree structure outperform both implicit reasoning models and local heuristics on both solve rate and search efficiency.
\end{itemize}

\section{Related Work}

\paragraph{Chain-of-thought Reasoning and Self-reflection.}
Chain-of-thought \citep{wei2022chain, nye2021show} prompting improves LLM reasoning by eliciting a sequence of natural-language reasoning steps before the final answer. Subsequent work extends this basic paradigm through problem decomposition, reflection, and iterative refinement, allowing models to try multiple approaches, break problems into simpler subproblems, or revise earlier attempts \citep{renze2024self, zhou2022least, madaan2023self, shinn2023reflexion}. These approaches make reasoning more flexible, but the branching, revision, and selection process is typically left implicit in the generated text rather than represented as an explicit search structure.

\paragraph{LLM Reasoning with External Controllers.}
A closely related line of work makes the search structure explicit by placing external controllers around the LLM. Tree-of-thought \citep{yao2023tree} uses an external search controller to expand, evaluate, and prune a tree of LLM-generated intermediate thoughts, enabling deliberate exploration and backtracking at inference time. LLM-A* \citep{meng2024llm} uses a prompted LLM to propose intermediate waypoints, while an external A*-based controller maintains the search frontier, verifies path validity, and constructs the final path. Graph of Thoughts \citep{besta2024graph} uses an external controller to organize LLM-generated intermediate thoughts into a graph, and uses a prompter module to select thoughts that go into the prompt. LATS \citep{zhou2023language} and Reasoning-as-Planning \citep{hao2023reasoning} wrap a pretrained LLM with an external Monte Carlo Tree Search controller, maintaining the full search tree externally while each language-model call conditions on the path that leads to the selected node. Self-Evaluation Guided Beam Search \citep{xie2023self} uses an external stochastic beam-search decoder to maintain and prune partial reasoning chains, where an LLM both proposes candidate reasoning steps and self-evaluates their stepwise correctness conditioned on a single reasoning path. These methods all use an external controller to maintain the search tree, while the LLM is only queried on a local view (the current path) rather than on the entire search trace. Our work compares this style of local-state LLM use against a trace-conditioned reasoning model, studying the effect of having access to the full search trace.

\section{Reasoning Environments}\label{sec:envs}

We evaluate on three fully observable reasoning domains: Blocks World \citep{valmeekam2022large}, Grid Navigation \citep{mittal2025learning}, and Sokoban \citep{boxobanlevels}. All three domains naturally require search, and can benefit from strong heuristics to improve search efficiency. For each domain, we generate 20k instances for SFT training, a separate 20k instances for RL, and 1k for validation. SFT traces are generated using best-first search with a rule-based heuristic for each domain, and instances whose SFT trace exceeds 16k tokens are filtered out. We describe each domain in detail below, including the set of available actions, state representations, and instance generation procedure.

\paragraph{Blocks World.} Blocks world is a reasoning problem in which the agent must transform an initial configuration of stacked blocks into a target stacking by moving a block with nothing on top of it. We generate instances with three stacks and between 4 and 10 blocks. The ratio of instances with different numbers of blocks in the training and testing datasets is weighted by the square of the number of blocks. The state is serialized as an ordered list of stacks, e.g., ``\texttt{S\{ 0:A<B ; 1:C ; 2:D \}}'', where each stack is written from bottom to top. An action moves one clear block from its current support to another stack or to the table, written as ``\texttt{B:A->TABLE}'' (moving B from the top of A to TABLE). The instances are generated by randomly sampling initial and goal states and filtering out instances that can be solved in less than 4 steps. We generate search traces for training with best-first search algorithm with a rule-based heuristic. The heuristic counts the number of ``good'' blocks in a state, where a block is considered ``good'' if it is already on its goal support and every block beneath it is also ``good.''

\paragraph{Grid Navigation.}
The navigation domain asks the agent to move on a \(10\times 10\) 2D grid from a randomly selected starting cell to a randomly selected goal cell while avoiding obstacles. Each grid has 35\% probability of being an obstacle. We generate instances by randomly sampling obstacles in the map and retain only instances whose optimal path length is at least 6 and whose number of best-first expansions is at least 1.3 times the resulting plan length. When using the LLM reasoning model, the full grid is given in the input prompt, and during search the state records only the agent location, written as ``\texttt{A=(r,c)}''. The action space is the four cardinal moves \texttt{U}, \texttt{D}, \texttt{L}, and \texttt{R}. We use best-first search to produce the SFT traces, with Manhattan distance to the goal as the heuristic. This heuristic can be misleading in this domain because minimizing Manhattan distance can easily lead to dead ends, leaving room for the model to learn a stronger search heuristic.

\paragraph{Sokoban.}
Sokoban is a classic game in which the player navigates a 2D grid and pushes boxes onto goal squares. We use the unfiltered split from Boxoban levels \citep{boxobanlevels} for training and testing. The original Boxoban instances have four boxes each, for which the search can have more than 5000 expansions. To make search manageable for the LLM, we randomly remove two boxes and goal positions in each instance. The full grid configuration, including the initial box, wall, and goal locations, is given in the prompt. During search, the state is represented by the player position together with the set of box positions, written as ``\texttt{P=(r,c) B=\{(r,c),\ldots\}}''. The action space is again \texttt{U}, \texttt{D}, \texttt{L}, and \texttt{R}, which moves the player. We again generate traces with best-first search. The heuristic \citep{junghanns1997sokoban} first estimates, for each box-goal pair, the number of moves required to push the box to the goal while ignoring the other boxes, and then uses the minimum bipartite matching cost between boxes and goals as the heuristic value. Sokoban is the hardest domain among the three, with longer search traces and more complex transition dynamics, providing a challenging testbed for different methods.

\section{Effect of Search Trace Access}\label{sec:history}

We begin by studying whether having access to the search trace makes an LLM a better search agent. To do so, we compare a trace-conditioned reasoning policy against a local-state search policy. In the trace-conditioned setting, the LLM observes the full search trace and directly predicts the next search step. In the local-state setting, we run an external best-first search algorithm guided by a heuristic, learned using the same LLM, that observes only the current candidate and the goal when expanding.

\subsection{Trace-conditioned LLM reasoning policy}\label{sec:reasoning-training}

We train an LLM reasoning policy that conditions on the full search history and predicts the next search step. The search trace is linearized as a sequence of expansions, with each search step written using the template
\begin{quote}
\texttt{EXPAND ACT \{action\} -> \{resulting state\}}
\end{quote}
This serialization mirrors the style of current reasoning traces, where the model tries different actions and backtracks when a branch fails but does not explicitly label the tree structure. Nevertheless, conditioning on the preceding trace allows the policy, in principle, to use non-local context to guide search decisions.

We train the LLM reasoning policy with a two-stage pipeline: supervised fine-tuning (SFT) on best-first search traces, followed by reinforcement learning with GRPO. We describe the training details as follows.

\paragraph{Stage 1: SFT on best-first traces.}
For each domain, we generate search traces by running best-first search with a rule-based heuristic as described in Section \ref{sec:envs}. The resulting traces demonstrate valid search behavior (branch expansion, backtracking, and solution construction) and serve as the SFT data for the policy. Each SFT example consists of a prompt and a completion. The prompt contains the instance configuration: the initial and goal states for Blocks World, the grid with obstacles together with the start and goal cells for Navigation, and the grid with walls, boxes, and goals for Sokoban. The completion has two parts: (i) the thinking process, consisting of the search trace, written as a sequence of expansions, and (ii) the final plan, which is the sequence of actions that leads to the goal extracted from the search trace. The model is trained to generate both parts, conditioned on the prompt. We use Qwen3-0.6B \citep{yang2025qwen3} as the backbone LLM and train for 4 epochs with learning rate 1e-5.

\paragraph{Stage 2: GRPO for search efficiency.}
After SFT, the LLM is initialized with reasonable search behavior, but since the rule-based heuristic that produced the SFT traces may not be optimal and the model may not yet fully exploit its access to the trace, we further optimize the policy with GRPO \citep{shao2024deepseekmath}, with a reward that encourages the model to improve its search for both correctness and efficiency.

The reward has two components. First, the model receives a correctness reward that is either 1 or 0, depending on whether the generated search trace is valid and the resulting plan correctly solves the task. Second, when the trace and plan are correct, we apply an additional efficiency term that penalizes longer searches:
\begin{equation}
\label{eq:reward}
R(\tau) = \mathbf{1}[\mathrm{valid}(\tau) \land \mathrm{correct}(\tau)]\left(1 - \lambda \sum_{t=0}^{N_{\mathrm{exp}}(\tau)-1} \gamma^t\right),
\end{equation}
where \(N_{\mathrm{exp}}(\tau)\) is the number of search expansions in trace \(\tau\), \(0 < \gamma < 1\) is the geometric decay factor, and \(\lambda\) sets the penalty scale. Each additional expansion subtracts one more discounted term from the correctness reward, so shorter correct traces receive higher return. Since the efficiency penalty is a finite geometric sum, it remains bounded; choosing \(\lambda < 1-\gamma\) ensures that every correct trace still receives positive reward and therefore scores above any incorrect trace. In Appendix~\ref{app:reward-alternatives} we also discuss some alternative reward function designs. In our experiments we set $\lambda = 0.005$ and $\gamma = 0.99$. GRPO samples $N = 8$ rollouts per instance for the advantage estimate, and is run for 1 epoch with learning rate 1e-5. All training runs in this paper are performed on one NVIDIA H100 GPU.

\subsection{Local-state LLM heuristics}
Our local-state LLM heuristic resembles a common pattern for applying LLMs to search problems: the LLM is invoked locally to score or propose continuations from a single tree node, while an outer search procedure decides which node to expand next. Tree-of-thoughts \citep{yao2023tree} is one instance of this pattern: the LLM scores candidate ``thoughts'' branching from a single node while the surrounding controller maintains the search tree. One feature of this paradigm is that the LLM is queried on a \emph{local} view (the current path) rather than on the entire search trace. Comparing this style of local-state LLM use against a trace-conditioned reasoning model, we study the effect of having access to the full search trace. Note that tree-of-thoughts typically conditions on the full path from the root to the current node, whereas our heuristic conditions only on the leaf state and the goal (examples of per-domain inputs are listed in Table~\ref{tab:heuristic-training}). This is fine in our setting since all three environments are Markovian, and thus the current state and goal are sufficient for the heuristic to evaluate candidate actions with respect to future progress toward the goal.

\paragraph{LLM heuristic-guided tree search.}
We run best-first search guided by the LLM heuristic. At each expansion step, for each (state, action) candidate on the frontier of leaf nodes, we query the LLM heuristic with the current state, the candidate action, and the goal, and obtain a heuristic value via an additional MLP head. The best-first algorithm picks the candidate $(s^\star, a^\star)$ with the lowest heuristic value, applies the action to obtain the resulting state, and adds it as a child of $s^\star$ in the search tree. This continues until a solution is found or the search budget is exhausted. The full procedure is given in Algorithm~\ref{alg:bfs}. We set the search budget $B$ to 500 expansions during RL training and 200 during evaluation. The original tree-of-thoughts is a training-free inference-time method, but for a fair comparison with the trace-conditioned reasoning policy of Section~\ref{sec:reasoning-training}, we train the LLM heuristic with an analogous two-stage pipeline.

\begin{algorithm}[t]
\caption{LLM heuristic-guided best-first tree search.}
\label{alg:bfs}
\begin{algorithmic}[1]
\Require initial state $s_0$, goal $g$, transition function $T$, LLM heuristic $h_\theta(s, a, g)$, search budget $B$
\Ensure plan from $s_0$ to a goal state, or \textsc{fail}
\State Initialize the search tree with root $s_0$
\State $\mathcal{V} \gets \{s_0\}$ \Comment{visited states, used for cycle avoidance}
\State $\mathcal{F} \gets \{(s_0, a) : a \in \mathrm{Actions}(s_0)\}$ \Comment{frontier of (leaf-state, candidate-action) pairs}
\For{$t = 1$ to $B$}
    \For{each $(s, a) \in \mathcal{F}$}
        \State Compute $h_\theta(s, a, g)$ \Comment{LLM heuristic scores candidate state-action pairs}
    \EndFor
    \State $(s^\star, a^\star) \gets$ candidate with the lowest $h_\theta$ value \Comment{or sample from softmax during training}
    \State $s' \gets T(s^\star, a^\star)$ \Comment{environment transition for the chosen action}
    \State Add $s'$ as a child of $s^\star$ in the tree;\ \ $\mathcal{V} \gets \mathcal{V} \cup \{s'\}$
    \If{$s'$ satisfies $g$}
        \State \Return the path from $s_0$ to $s'$ in the tree
    \EndIf
    \State $\mathcal{F} \gets \big(\mathcal{F} \setminus \{(s^\star, a^\star)\}\big) \cup \{(s', a) : a \in \mathrm{Actions}(s'),\, T(s', a) \notin \mathcal{V}\}$ \Comment{skip already-visited successors}
\EndFor
\State \Return \textsc{fail}
\end{algorithmic}
\end{algorithm}

\paragraph{Supervised fine-tuning.} The LLM backbone with its MLP head is trained on (state, action) pairs collected from the SFT traces, with the rule-based heuristic value as the target. This mirrors Stage 1 in Section~\ref{sec:reasoning-training} and initializes the heuristic for subsequent reinforcement learning.

\paragraph{Reinforcement learning.} We formulate the tree expansion process of selecting which candidate to expand as a decision-making problem \citep{mittal2025learning}, guided by the LLM heuristic. We can then optimize the LLM heuristic with reinforcement learning, using the same reward (Eq.~\ref{eq:reward}) used for the trace-conditioned reasoning policy. Since the search uses the oracle transition function and goal verifier, every trace is valid; therefore, the correctness term is zero only when the search fails to find a goal within the budget. We use the same advantage calculation as in GRPO. For each instance, we run search $N$ times, compute reward $r_i$ for each trace, and use the standardized advantage
\begin{equation}
\label{eq:advantage}
A_i = \dfrac{r_i - \mu}{\sigma + \epsilon},
\end{equation}
where $\mu$ and $\sigma$ are the mean and standard deviation of the rewards across the $N$ traces. The resulting policy gradient is averaged over the $N$ traces sampled from the tree expansion policy $\pi_\theta$ induced by the LLM heuristic:
\begin{equation}
\label{eq:policy-grad}
\nabla_\theta J(\theta) = \frac{1}{N} \sum_{i=1}^{N} A_i \sum_{t=0}^{T_i - 1} \nabla_\theta \log \pi_\theta\!\left(c_{i,t} \mid \mathcal{C}_{i,t}\right),
\end{equation}
where $\mathcal{C}_{i,t}$ is the set of candidate expansions on the frontier at step $t$ of trace $i$, $c_{i,t} \in \mathcal{C}_{i,t}$ is the candidate sampled and expanded at that step, $T_i$ is the number of expansions in trace $i$, and the per-step distribution is the softmax over negated LLM heuristic scores, since lower heuristics are favored, 
\begin{equation}
\label{eq:cand-softmax}
\pi_\theta(c \mid \mathcal{C}_{i,t}) = \dfrac{\exp\!\left(-h_\theta(c)\right)}{\sum_{c' \in \mathcal{C}_{i,t}} \exp\!\left(-h_\theta(c')\right)}.
\end{equation}
We train the LLM heuristic for 1 epoch of SFT and then 1 epoch of RL, using the same SFT and RL training instances as the reasoning policy.

\begin{table}[t]
    \caption{Per-domain training data for the local-state LLM heuristic. Each training pair consists of a (state, action) input collected from the SFT search traces and a target value computed by the rule-based heuristic.}
    \label{tab:heuristic-training}
    \centering
    \small
    \renewcommand{\arraystretch}{1.25}
    \begin{tabularx}{\linewidth}{l c l X}
        \toprule
        Domain & \# SFT Pairs & Target heuristic & Example input \\
        \midrule
        Blocks World & 1.74M & Number of ``good'' blocks &
            goal \texttt{S\{0:D<C<B<A\}}, state \texttt{S\{0:A<B; 1:C; 2:D\}}, \newline action \texttt{B:A->TABLE} \\
        Navigation & 1.22M & Manhattan distance to goal &
            \{grid\}, goal \texttt{(7,8)}, state \texttt{P=(3,2)}, \newline action \texttt{R} \\
        Sokoban & 1.28M & Min-cost box-to-goal matching &
            \{grid\}, goal \texttt{\{(7,7),(1,1)\}}, state \texttt{P=(2,3) B=\{(4,5),(6,2)\}}, \newline
            action \texttt{U} \\
        \bottomrule
    \end{tabularx}
\end{table}

\begin{table*}[t]
    \caption{Policies with versus without access to the search trace, evaluated on the three reasoning domains. Lower is better for expansions. Subscripts show standard errors.}
    \label{tab:history-vs-local}
    \centering
    {\small
    \renewcommand{\arraystretch}{1.15}
    \begin{tabular*}{\textwidth}{@{\extracolsep{\fill}}lcccccc@{}}
        \toprule
        & \multicolumn{2}{c}{Blocks World} & \multicolumn{2}{c}{Navigation} & \multicolumn{2}{c}{Sokoban} \\
        \cmidrule(r){2-3} \cmidrule(r){4-5} \cmidrule(r){6-7}
        Method & Success $\uparrow$ & Expansions $\downarrow$ & Success $\uparrow$ & Expansions $\downarrow$ & Success $\uparrow$ & Expansions $\downarrow$ \\
        \midrule
        BFS (Pretrained) & $99.7_{\pm 0.17}$ & $9.54_{\pm 0.31}$ & $100.0_{\pm 0.00}$ & $20.27_{\pm 0.42}$ & $94.9_{\pm 0.70}$ & $73.21_{\pm 2.11}$ \\
        BFS (RL)         & $99.8_{\pm 0.14}$ & $8.56_{\pm 0.18}$ & $100.0_{\pm 0.00}$ & $16.99_{\pm 0.37}$ & $99.1_{\pm 0.30}$ & $64.08_{\pm 1.86}$ \\
        SFT-implicit     & $90.0_{\pm 0.95}$ & $9.12_{\pm 0.16}$ & $90.6_{\pm 0.92}$  & $15.39_{\pm 0.18}$ & $74.8_{\pm 1.37}$ & $68.24_{\pm 7.70}$ \\
        GRPO-implicit    & $97.3_{\pm 0.51}$ & $8.25_{\pm 0.14}$ & $94.9_{\pm 0.70}$  & $14.80_{\pm 0.16}$ & $85.9_{\pm 1.10}$ & $63.54_{\pm 4.12}$ \\
        \bottomrule
    \end{tabular*}
    }
\end{table*}

\subsection{Results}

Table~\ref{tab:history-vs-local} shows that the implicit reasoning model achieves lower success rates than BFS (RL), while offering only modest improvements in search expansions. Although the implicit reasoning policy conditions on the whole trace, it is not reliably stronger than the local-state heuristic baseline. GRPO improves the implicit policy over its SFT initialization, but this alone still does not establish a clear advantage over local-state heuristic-guided search.

One explanation is that the available trace is simply a flat serialization of a search tree. The model sees the history, but the structure of the tree is lost in the serialization. This leads to the next question: would making the search tree structure explicit allow the model to better use the available history and therefore outperform local-state heuristics? We investigate this question in the next section by comparing implicit traces against a more structured trace representation that explicitly labels the parent-child relationships in the search tree.

\section{Structured Search Traces}\label{sec:pipeline}

In this section, we investigate whether making the search history of LLM reasoning models more structured can make the access to such histories more useful. To do that we train a second policy on the same underlying search traces but with explicit parent-pointer annotations that expose the tree structure. We then compare this explicit-trace policy against the implicit-trace policy of Section~\ref{sec:reasoning-training} and the local-state heuristic baseline of Section~\ref{sec:history}.

\subsection{Trace representations}\label{sec:trace-representations}

To make the search trace more useful, we introduce a simple modification to make it a more structured representation, where each expansion identifies the frontier state being expanded and the child state reached by the chosen action. We refer to this source-state identifier as a \emph{parent pointer}, because it tells the model exactly which earlier search state the new expansion descends from. In contrast, the implicit representation in Section~\ref{sec:reasoning-training} records only the action and resulting state, leaving the resumed search state to be inferred from context. The two representations contain the same search behavior but differ in whether the tree topology is directly exposed.

An example of the structured representation is shown below:
\begin{quote}
\texttt{EXPAND }\textcolor{green!50!black}{\texttt{sid=0}}\texttt{ ACT \{action\} -> }\textcolor{green!50!black}{\texttt{sid=1}}\texttt{ \{resulting state\}}\\
\texttt{EXPAND }\textcolor{green!50!black}{\texttt{sid=1}}\texttt{ ACT \{action\} -> }\textcolor{green!50!black}{\texttt{sid=2}}\texttt{ \{resulting state\}}\\
\texttt{EXPAND }\textcolor{green!50!black}{\texttt{sid=1}}\texttt{ ACT \{action\} -> }\textcolor{green!50!black}{\texttt{sid=3}}\texttt{ \{resulting state\}}
\end{quote}
Full per-domain SFT examples for both the implicit and explicit formats are provided in Appendix~\ref{app:sft-examples}, and a visualization for each domain is shown in Appendix Figure~\ref{fig:domain-examples}.

The explicit policy is trained on the same instances using the SFT+GRPO pipeline as the implicit policy of Section~\ref{sec:reasoning-training}, with the base model and training parameters held fixed. Comparing the two policies trained on the same underlying searches but with different trace representations isolates the effect of making the tree structure explicit.

\subsection{Effect of structured trace annotations}

\begin{table}[t]
\centering
\caption{Implicit versus explicit trace annotations on controlled reasoning domains. Lower is better for expansions. Subscripts show standard errors computed over evaluation instances.}
\label{tab:main-results}
\begin{tabular}{llrr}
\toprule
\multicolumn{4}{c}{\textbf{Blocks World}} \\
\midrule
Method & Training & Success $\uparrow$ & Expansions $\downarrow$ \\
\midrule
No parent pointer & SFT & $90.0_{\pm 0.95}$ & $9.12_{\pm 0.16}$ \\
With parent pointer & SFT & $89.6_{\pm 0.97}$ & $9.06_{\pm 0.45}$ \\
No parent pointer & GRPO & $97.3_{\pm 0.51}$ & $8.25_{\pm 0.14}$ \\
With parent pointer & GRPO & $100.0_{\pm 0.00}$ & $7.31_{\pm 0.08}$ \\
-- & BFS (pretrained) & $99.7_{\pm 0.17}$ & $9.54_{\pm 0.31}$ \\
-- & BFS (RL) & $99.8_{\pm 0.14}$ & $8.56_{\pm 0.18}$ \\
\midrule
\multicolumn{4}{c}{\textbf{Sokoban}} \\
\midrule
Method & Training & Success $\uparrow$ & Expansions $\downarrow$ \\
\midrule
No parent pointer & SFT & $74.8_{\pm 1.37}$ & $68.24_{\pm 7.70}$ \\
With parent pointer & SFT & $85.6_{\pm 1.11}$ & $73.26_{\pm 5.88}$ \\
No parent pointer & GRPO & $85.9_{\pm 1.10}$ & $63.54_{\pm 4.12}$ \\
With parent pointer & GRPO & $89.6_{\pm 0.97}$ & $52.82_{\pm 3.35}$ \\
With parent pointer & GRPO + generation constraint & $98.9_{\pm 0.33}$ & $54.70_{\pm 4.62}$ \\
-- & BFS (pretrained) & $94.9_{\pm 0.70}$ & $73.21_{\pm 2.11}$ \\
-- & BFS (RL) & $99.1_{\pm 0.30}$ & $64.08_{\pm 1.86}$ \\
\midrule
\multicolumn{4}{c}{\textbf{Navigation}} \\
\midrule
Method & Training & Success $\uparrow$ & Expansions $\downarrow$ \\
\midrule
No parent pointer & SFT & $90.6_{\pm 0.92}$ & $15.39_{\pm 0.18}$ \\
With parent pointer & SFT & $95.2_{\pm 0.68}$ & $15.63_{\pm 0.18}$ \\
No parent pointer & GRPO & $94.9_{\pm 0.70}$ & $14.80_{\pm 0.16}$ \\
With parent pointer & GRPO & $100.0_{\pm 0.00}$ & $14.28_{\pm 0.12}$ \\
-- & BFS (pretrained) & $100.0_{\pm 0.00}$ & $20.27_{\pm 0.42}$ \\
-- & BFS (RL) & $100.0_{\pm 0.00}$ & $16.99_{\pm 0.37}$ \\
\bottomrule
\end{tabular}
\end{table}

We first compare the effects of the two different trace representations by comparing four policies in this pipeline: SFT-implicit, SFT-explicit, GRPO-implicit, and GRPO-explicit. At the SFT stage, explicit annotation raises average solve rate by 5.0 points across domains, driven mainly by gains in Navigation (95.2 vs.~90.6) and Sokoban (85.6 vs.~74.8), while Blocks World is essentially unchanged (89.6 vs.~90.0). The effect on search efficiency before RL is weaker: the explicit model uses nearly the same number of expansions in Blocks World (9.06 vs.~9.12), but slightly more in Navigation and Sokoban. This suggests that explicit traces primarily improve the model's ability to recover valid search behavior during imitation.

We then initialize GRPO from the SFT checkpoints. After GRPO training, the explicit model consistently outperforms the implicit model. GRPO-explicit reaches 100.0\% solve rate in both Blocks World and Navigation and 89.6\% in Sokoban, compared with 97.3\%, 94.9\%, and 85.9\% for GRPO-implicit. It is also more search-efficient in every domain, reducing expansions from 8.25 to 7.31 in Blocks World, from 14.80 to 14.28 in Navigation, and from 63.54 to 52.82 in Sokoban. Note that in Sokoban, the explicit model's solve rate is still below the local-state heuristic, this may be attributed to the harder transition dynamics of the domain, where an action may cause both the player position and box positions to change, and since there are more search steps, the LLM reasoning model with free-form generation may be more prone to compounding errors. In contrast, the BFS with local heuristic leverages an external transition function to produce the next state given the action, which is always correct. For a fair comparison, during evaluation, we apply a generation constraint that prevents the model from generating invalid actions (e.g., moving into a wall or pushing a box into an occupied cell) to match the local-state heuristic's access to the environment's transition function. Constraining the LLM output to follow specific constraints is commonly done in practice, see e.g. \cite{dong2025xgrammar}. Note that this generation constraint is only applicable to the explicit model since we need to know the state the action is applied to. With this constraint, the explicit model gets a 98.9\% solve rate in Sokoban, and matches the local-state heuristic's solve rate of 99.1\% while still using fewer expansions (54.70 vs.~64.08).

Overall, explicit annotation improves solve rate at the SFT stage and improves both solve rate and search efficiency after GRPO. This pattern suggests that making the search tree structure explicit not only makes demonstrated search behavior easier to learn during SFT, but also provides a stronger starting point for subsequent policy optimization. A likely reason is that explicit parent pointers help the model track the search tree and the history of visited states more reliably, which in turn enables more effective backtracking and search decisions during RL. Taken together with the comparison in Section~\ref{sec:history}, these results suggest that the benefits of search trace access are most fully realized when the tree structure of the trace is made explicit, motivating search-structured supervision as a simple way to improve LLM reasoning.

\subsection{Analysis of Different Policies}

\paragraph{Explicit structure helps extracting the plan from search trace.}
One potential benefit of explicit parent annotations is that it makes it easier for the model to extract the final plan from its own search trace. To quantify this, we look at the cases where the model fails to generate a correct plan and ask how many of them are extraction failures rather than search failures: the search actually found a correct plan that leads to the goal, but the model generates an incorrect plan. For the implicit policy, since the tree structure is not directly available from the search trace, we determine whether the search found a plan by checking whether a path from the initial state to the goal state can be constructed using the states visited in the search.

Table~\ref{tab:extraction-failure} reports this breakdown for both training stages. The results show a clear pattern: conditioned on the search trace finding a plan, the explicit policy has fewer extraction failures. This suggests that the search history with explicit tree structure annotation allows the model to more reliably extract the correct plan from the search trace, which may be one reason why it outperforms the implicit policy.

\paragraph{Explicit trace-conditioned LLM explores state space more effectively.}
We study the search behavior of different policies by looking at the distribution of visited states during search in the navigation domain, where we define the distance between two states as the shortest-path distance between them on the grid. Given a search tree, we measure the diversity of the set of visited states $S$ by the average pairwise distance \citep{rao1982diversity},
\begin{equation}
\label{eq:avg-dist}
\bar{d}(S) = \frac{1}{|S|^2} \sum_{s, s' \in S} d(s, s'),
\end{equation}
where $d(s, s')$ is the shortest-path distance between $s$ and $s'$. Higher values indicate that the visited states are spread further apart on the grid.

Table~\ref{tab:avg-dist} reports the average $\bar{d}$ across navigation instances. The explicit policy attains the highest value (4.19), followed by the implicit policy (4.11) and the local-state heuristic (3.99). A higher $\bar{d}$ means the visited states are spread further apart on the grid, while a lower value indicates that the policy concentrates its expansions in a small neighborhood. The trace-conditioned policies, especially the explicit policy, thus cover more of the state space in its search. A plausible explanation is that conditioning on the search trace lets the policy see which neighborhoods earlier expansions already covered and steer subsequent expansions elsewhere, whereas the local-state heuristic has no memory of past expansions to guide such avoidance; explicit parent pointers further amplify this effect by making the explored region of the tree easier to track.

\begin{table}[t]
\centering
\begin{minipage}[t]{0.58\textwidth}
\centering
\caption{Plan-extraction failures among incorrect runs. Each cell reports the fraction of failed cases in which the failing reason is extraction failure. ``N/A'' indicates that the policy has no failed cases on that domain (100\% solve rate).}
\label{tab:extraction-failure}
\small
\renewcommand{\arraystretch}{1.15}
\begin{tabular}{lccc}
    \toprule
    Method & Blocks World & Navigation & Sokoban \\
    \midrule
    SFT-implicit  & 41.44\% & 96.67\% & 80.78\% \\
    SFT-explicit  & 26.00\% & 94.52\% & 54.17\% \\
    GRPO-implicit & 74.07\% & 94.12\% & 71.43\% \\
    GRPO-explicit & N/A     & N/A     & 45.46\% \\
    \bottomrule
\end{tabular}
\end{minipage}%
\hfill
\begin{minipage}[t]{0.40\textwidth}
\centering
\caption{Average pairwise distance $\bar{d}$ of visited states in the search trees produced by different policies on Navigation. Higher values indicate more diverse exploration of the state space.}
\vspace{+1mm}
\label{tab:avg-dist}
\small
\renewcommand{\arraystretch}{1.15}
\begin{tabular}{lc}
    \toprule
    Method & $\bar{d}$ $\uparrow$ \\
    \midrule
    BFS (RL)      & 3.99 \\
    GRPO-implicit & 4.11 \\
    GRPO-explicit & 4.19 \\
    \bottomrule
\end{tabular}
\end{minipage}
\end{table}

\section{Conclusion}

In this paper, we studied LLM reasoning from a search perspective, asking whether access to the full search trace makes a reasoning model a stronger search policy than a local-state LLM heuristic. Across Blocks World, Navigation, and Sokoban, we found that trace access alone does not reliably yield better search: when the underlying tree structure is left implicit, the model often fails to fully exploit the additional history. We then showed that a minimal change to the trace representation, adding parent pointers that explicitly identify which search state each expansion descends from, improves both solve rate and search efficiency. Further analysis suggests that explicit structure helps the model recover final plans from its own traces and explore the state space more effectively.

These results indicate that the benefit of trace-conditioned reasoning depends not only on whether the model sees the search context, but also on how that context is represented. In particular, making the search tree explicit turns an unstructured history of attempted steps into a representation that is easier to track, optimize, and use for plan extraction. More broadly, our findings suggest that improving LLM reasoning may require not only better training objectives or larger models, but also better interfaces between reasoning traces and the underlying computational structures they represent. Search-structured supervision offers a simple and promising direction for making LLM reasoning more efficient and easier to optimize.

\paragraph{Limitations} Our experiments are limited to three fully observable, controlled search domains, and additional designs are needed to transfer the findings to open-ended reasoning tasks. In addition, due to limited computational resources, the experiments are conducted on one base LLM, leaving open how the benefits scale with larger models.

\bibliographystyle{plainnat}
\bibliography{references}

\begin{thebibliography}{31}
\providecommand{\natexlab}[1]{#1}
\providecommand{\url}[1]{\texttt{#1}}
\expandafter\ifx\csname urlstyle\endcsname\relax
  \providecommand{\doi}[1]{doi: #1}\else
  \providecommand{\doi}{doi: \begingroup \urlstyle{rm}\Url}\fi

\bibitem[Besta et~al.(2024)Besta, Blach, Kubicek, Gerstenberger, Podstawski, Gianinazzi, Gajda, Lehmann, Niewiadomski, Nyczyk, et~al.]{besta2024graph}
Maciej Besta, Nils Blach, Ales Kubicek, Robert Gerstenberger, Michal Podstawski, Lukas Gianinazzi, Joanna Gajda, Tomasz Lehmann, Hubert Niewiadomski, Piotr Nyczyk, et~al.
\newblock Graph of thoughts: Solving elaborate problems with large language models.
\newblock In \emph{Proceedings of the AAAI conference on artificial intelligence}, volume~38, pages 17682--17690, 2024.

\bibitem[Chen and Li(2024)]{chen2024toward}
Sijia Chen and Baochun Li.
\newblock Toward adaptive reasoning in large language models with thought rollback.
\newblock \emph{arXiv preprint arXiv:2412.19707}, 2024.

\bibitem[Chen et~al.(2024)Chen, Li, and Niu]{chen2024boosting}
Sijia Chen, Baochun Li, and Di~Niu.
\newblock Boosting of thoughts: Trial-and-error problem solving with large language models.
\newblock \emph{arXiv preprint arXiv:2402.11140}, 2024.

\bibitem[Dong et~al.(2025)Dong, Ruan, Cai, Xu, Zhao, Lai, and Chen]{dong2025xgrammar}
Yixin Dong, Charlie~F Ruan, Yaxing Cai, Ziyi Xu, Yilong Zhao, Ruihang Lai, and Tianqi Chen.
\newblock Xgrammar: Flexible and efficient structured generation engine for large language models.
\newblock \emph{Proceedings of Machine Learning and Systems}, 7, 2025.

\bibitem[Guez et~al.(2018)Guez, Mirza, Gregor, Kabra, Racaniere, Weber, Raposo, Santoro, Orseau, Eccles, Wayne, Silver, Lillicrap, and Valdes]{boxobanlevels}
Arthur Guez, Mehdi Mirza, Karol Gregor, Rishabh Kabra, Sebastien Racaniere, Theophane Weber, David Raposo, Adam Santoro, Laurent Orseau, Tom Eccles, Greg Wayne, David Silver, Timothy Lillicrap, and Victor Valdes.
\newblock An investigation of model-free planning: boxoban levels.
\newblock https://github.com/deepmind/boxoban-levels/, 2018.

\bibitem[Guo et~al.(2025)Guo, Yang, Zhang, Song, Wang, Zhu, Xu, Zhang, Ma, Bi, et~al.]{guo2025deepseek}
Daya Guo, Dejian Yang, Haowei Zhang, Junxiao Song, Peiyi Wang, Qihao Zhu, Runxin Xu, Ruoyu Zhang, Shirong Ma, Xiao Bi, et~al.
\newblock Deepseek-r1: Incentivizing reasoning capability in llms via reinforcement learning.
\newblock \emph{arXiv preprint arXiv:2501.12948}, 2025.

\bibitem[Hao et~al.(2023)Hao, Gu, Ma, Hong, Wang, Wang, and Hu]{hao2023reasoning}
Shibo Hao, Yi~Gu, Haodi Ma, Joshua Hong, Zhen Wang, Daisy Wang, and Zhiting Hu.
\newblock Reasoning with language model is planning with world model.
\newblock In \emph{Proceedings of the 2023 Conference on Empirical Methods in Natural Language Processing}, pages 8154--8173, 2023.

\bibitem[Hart et~al.(1968)Hart, Nilsson, and Raphael]{hart1968formal}
Peter~E Hart, Nils~J Nilsson, and Bertram Raphael.
\newblock A formal basis for the heuristic determination of minimum cost paths.
\newblock \emph{IEEE transactions on Systems Science and Cybernetics}, 4\penalty0 (2):\penalty0 100--107, 1968.

\bibitem[Jaech et~al.(2024)Jaech, Kalai, Lerer, Richardson, El-Kishky, Low, Helyar, Madry, Beutel, Carney, et~al.]{jaech2024openai}
Aaron Jaech, Adam Kalai, Adam Lerer, Adam Richardson, Ahmed El-Kishky, Aiden Low, Alec Helyar, Aleksander Madry, Alex Beutel, Alex Carney, et~al.
\newblock Openai o1 system card.
\newblock \emph{arXiv preprint arXiv:2412.16720}, 2024.

\bibitem[Junghanns and Schaeffer(1997)]{junghanns1997sokoban}
Andreas Junghanns and Jonathan Schaeffer.
\newblock Sokoban: A challenging single-agent search problem.
\newblock In \emph{IJCAI Workshop on Using Games as an Experimental Testbed for AI Reasearch}, pages 27--36. Morgan Kaufmann Publishers San Francisco, CA, 1997.

\bibitem[Katz et~al.(2024)Katz, Kokel, Srinivas, and Sohrabi]{katz2024thought}
Michael Katz, Harsha Kokel, Kavitha Srinivas, and Shirin Sohrabi.
\newblock Thought of search: Planning with language models through the lens of efficiency.
\newblock \emph{Advances in Neural Information Processing Systems}, 37:\penalty0 138491--138568, 2024.

\bibitem[Li et~al.(2025)Li, Cao, Griggs, Liu, Mo, Tang, Hegde, Hakhamaneshi, Patil, Zaharia, et~al.]{li2025llms}
Dacheng Li, Shiyi Cao, Tyler Griggs, Shu Liu, Xiangxi Mo, Eric Tang, Sumanth Hegde, Kourosh Hakhamaneshi, Shishir~G Patil, Matei Zaharia, et~al.
\newblock Llms can easily learn to reason from demonstrations structure, not content, is what matters!
\newblock \emph{arXiv preprint arXiv:2502.07374}, 2025.

\bibitem[Madaan et~al.(2023)Madaan, Tandon, Gupta, Hallinan, Gao, Wiegreffe, Alon, Dziri, Prabhumoye, Yang, et~al.]{madaan2023self}
Aman Madaan, Niket Tandon, Prakhar Gupta, Skyler Hallinan, Luyu Gao, Sarah Wiegreffe, Uri Alon, Nouha Dziri, Shrimai Prabhumoye, Yiming Yang, et~al.
\newblock Self-refine: Iterative refinement with self-feedback.
\newblock \emph{Advances in neural information processing systems}, 36:\penalty0 46534--46594, 2023.

\bibitem[Meng et~al.(2024)Meng, Wang, Yang, Peng, and Chang]{meng2024llm}
Silin Meng, Yiwei Wang, Cheng-Fu Yang, Nanyun Peng, and Kai-Wei Chang.
\newblock Llm-a*: Large language model enhanced incremental heuristic search on path planning.
\newblock In \emph{Findings of the Association for Computational Linguistics: EMNLP 2024}, pages 1087--1102, 2024.

\bibitem[Mittal et~al.(2025)Mittal, Kang, and Lee]{mittal2025learning}
Dixant Mittal, Liwei Kang, and Wee~Sun Lee.
\newblock Learning to search from demonstration sequences.
\newblock In \emph{The Thirteenth International Conference on Learning Representations}, 2025.

\bibitem[Nye et~al.(2021)Nye, Andreassen, Gur-Ari, Michalewski, Austin, Bieber, Dohan, Lewkowycz, Bosma, Luan, et~al.]{nye2021show}
Maxwell Nye, Anders~Johan Andreassen, Guy Gur-Ari, Henryk Michalewski, Jacob Austin, David Bieber, David Dohan, Aitor Lewkowycz, Maarten Bosma, David Luan, et~al.
\newblock Show your work: Scratchpads for intermediate computation with language models.
\newblock 2021.

\bibitem[Pearl(1983)]{pearl1983heuristics}
Judea Pearl.
\newblock Heuristics: intelligent search strategies for computer problem solving.
\newblock 1983.

\bibitem[Rao(1982)]{rao1982diversity}
C~Radhakrishna Rao.
\newblock Diversity and dissimilarity coefficients: a unified approach.
\newblock \emph{Theoretical population biology}, 21\penalty0 (1):\penalty0 24--43, 1982.

\bibitem[Renze and Guven(2024)]{renze2024self}
Matthew Renze and Erhan Guven.
\newblock Self-reflection in llm agents: Effects on problem-solving performance.
\newblock \emph{arXiv preprint arXiv:2405.06682}, 2024.

\bibitem[Sel et~al.(2023)Sel, Al-Tawaha, Khattar, Jia, and Jin]{sel2023algorithm}
Bilgehan Sel, Ahmad Al-Tawaha, Vanshaj Khattar, Ruoxi Jia, and Ming Jin.
\newblock Algorithm of thoughts: Enhancing exploration of ideas in large language models.
\newblock \emph{arXiv preprint arXiv:2308.10379}, 2023.

\bibitem[Shao et~al.(2024)Shao, Wang, Zhu, Xu, Song, Bi, Zhang, Zhang, Li, Wu, et~al.]{shao2024deepseekmath}
Zhihong Shao, Peiyi Wang, Qihao Zhu, Runxin Xu, Junxiao Song, Xiao Bi, Haowei Zhang, Mingchuan Zhang, YK~Li, Yang Wu, et~al.
\newblock Deepseekmath: Pushing the limits of mathematical reasoning in open language models.
\newblock \emph{arXiv preprint arXiv:2402.03300}, 2024.

\bibitem[Shinn et~al.(2023)Shinn, Cassano, Gopinath, Narasimhan, and Yao]{shinn2023reflexion}
Noah Shinn, Federico Cassano, Ashwin Gopinath, Karthik Narasimhan, and Shunyu Yao.
\newblock Reflexion: Language agents with verbal reinforcement learning.
\newblock \emph{Advances in neural information processing systems}, 36:\penalty0 8634--8652, 2023.

\bibitem[Steinmetz and Hoffmann(2016)]{steinmetz2016towards}
Marcel Steinmetz and J{\"o}rg Hoffmann.
\newblock Towards clause-learning state space search: Learning to recognize dead-ends.
\newblock In \emph{Proceedings of the AAAI Conference on Artificial Intelligence}, volume~30, 2016.

\bibitem[Valmeekam et~al.(2022)Valmeekam, Olmo, Sreedharan, and Kambhampati]{valmeekam2022large}
Karthik Valmeekam, Alberto Olmo, Sarath Sreedharan, and Subbarao Kambhampati.
\newblock Large language models still can't plan (a benchmark for llms on planning and reasoning about change).
\newblock In \emph{NeurIPS 2022 Foundation Models for Decision Making Workshop}, 2022.

\bibitem[Wang et~al.(2025)Wang, Song, Tian, Yu, Mi, Duan, Tu, Su, and Yu]{wang2025don}
Ante Wang, Linfeng Song, Ye~Tian, Dian Yu, Haitao Mi, Xiangyu Duan, Zhaopeng Tu, Jinsong Su, and Dong Yu.
\newblock Don’t get lost in the trees: Streamlining llm reasoning by overcoming tree search exploration pitfalls.
\newblock In \emph{Proceedings of the 63rd Annual Meeting of the Association for Computational Linguistics (Volume 1: Long Papers)}, pages 23946--23959, 2025.

\bibitem[Wei et~al.(2022)Wei, Wang, Schuurmans, Bosma, Xia, Chi, Le, Zhou, et~al.]{wei2022chain}
Jason Wei, Xuezhi Wang, Dale Schuurmans, Maarten Bosma, Fei Xia, Ed~Chi, Quoc~V Le, Denny Zhou, et~al.
\newblock Chain-of-thought prompting elicits reasoning in large language models.
\newblock \emph{Advances in neural information processing systems}, 35:\penalty0 24824--24837, 2022.

\bibitem[Xie et~al.(2023)Xie, Kawaguchi, Zhao, Zhao, Kan, He, and Xie]{xie2023self}
Yuxi Xie, Kenji Kawaguchi, Yiran Zhao, James~Xu Zhao, Min-Yen Kan, Junxian He, and Michael Xie.
\newblock Self-evaluation guided beam search for reasoning.
\newblock \emph{Advances in Neural Information Processing Systems}, 36:\penalty0 41618--41650, 2023.

\bibitem[Yang et~al.(2025)Yang, Li, Yang, Zhang, Hui, Zheng, Yu, Gao, Huang, Lv, et~al.]{yang2025qwen3}
An~Yang, Anfeng Li, Baosong Yang, Beichen Zhang, Binyuan Hui, Bo~Zheng, Bowen Yu, Chang Gao, Chengen Huang, Chenxu Lv, et~al.
\newblock Qwen3 technical report.
\newblock \emph{arXiv preprint arXiv:2505.09388}, 2025.

\bibitem[Yao et~al.(2023)Yao, Yu, Zhao, Shafran, Griffiths, Cao, and Narasimhan]{yao2023tree}
Shunyu Yao, Dian Yu, Jeffrey Zhao, Izhak Shafran, Tom Griffiths, Yuan Cao, and Karthik Narasimhan.
\newblock Tree of thoughts: Deliberate problem solving with large language models.
\newblock \emph{Advances in neural information processing systems}, 36:\penalty0 11809--11822, 2023.

\bibitem[Zhou et~al.(2023)Zhou, Yan, Shlapentokh-Rothman, Wang, and Wang]{zhou2023language}
Andy Zhou, Kai Yan, Michal Shlapentokh-Rothman, Haohan Wang, and Yu-Xiong Wang.
\newblock Language agent tree search unifies reasoning acting and planning in language models.
\newblock \emph{arXiv preprint arXiv:2310.04406}, 2023.

\bibitem[Zhou et~al.(2022)Zhou, Sch{\"a}rli, Hou, Wei, Scales, Wang, Schuurmans, Cui, Bousquet, Le, et~al.]{zhou2022least}
Denny Zhou, Nathanael Sch{\"a}rli, Le~Hou, Jason Wei, Nathan Scales, Xuezhi Wang, Dale Schuurmans, Claire Cui, Olivier Bousquet, Quoc Le, et~al.
\newblock Least-to-most prompting enables complex reasoning in large language models.
\newblock \emph{arXiv preprint arXiv:2205.10625}, 2022.

\end{thebibliography}

\newpage
\appendix

\section{SFT Trace Examples}\label{app:sft-examples}

We show one short instance for each domain and trace formant, drawn from the training split. Within each domain, the implicit and explicit examples describe the same underlying instance and search but use different trace serializations.

Note that the Navigation traces explicitly include \texttt{BLOCKED} entries for actions that walk into walls, rather than silently pruning them. We found that without these entries, both the implicit and explicit models fail to learn a valid search policy during SFT. A plausible explanation is that the Manhattan-distance heuristic gives a strong directional preference over actions, so the model expects the highest-scoring candidate to be expanded next; if blocked candidates are silently removed from the trace, the visible expansion order no longer matches the heuristic's ranking, which may make the learning harder.

\subsection{Blocks World}

\noindent\textbf{Implicit format}
\begin{verbatim}
INSTANCE_BEGIN
BLOCKS A,B,C,D
INIT S{ 0:B<A<C ; 1:D }
GOAL S{ 0:B<A ; 1:D<C }
INSTANCE_END
SEARCH_TRACE_BEGIN
EXPAND S{ 0:B<A<C ; 1:D }
EXPAND ACT C:A->D -> S{ 0:B<A ; 1:D<C }
GOAL_REACHED
PLAN_LEN 1
PLAN_STEP 1 C:A->D
SEARCH_TRACE_END
\end{verbatim}

\noindent\textbf{Explicit format}
\begin{verbatim}
INSTANCE_BEGIN
BLOCKS A,B,C,D
INIT S{ 0:B<A<C ; 1:D }
GOAL S{ 0:B<A ; 1:D<C }
INSTANCE_END
SEARCH_TRACE_BEGIN
EXPAND sid=0 S{ 0:B<A<C ; 1:D }
EXPAND sid=0 ACT C:A->D -> sid=1 S{ 0:B<A ; 1:D<C }
GOAL_REACHED
PLAN_LEN 1
PLAN_STEP 1 C:A->D
SEARCH_TRACE_END
\end{verbatim}

\subsection{Navigation}

\noindent\textbf{Implicit format}
\begin{verbatim}
INSTANCE_BEGIN
GRID_SIZE 10
MAP
# # # # # # # # # #
# . . # # . # . . #
# . # . # . . . . #
# # # . . # . . . #
# # . . . . . # . #
# . G . . # . . . #
# . . . . . . . # #
# . # . . . S . # #
# . . . . . . # . #
# # # # # # # # # #
START (7,6)
GOAL (5,2)
INSTANCE_END
SEARCH_TRACE_BEGIN
EXPAND A=(7,6)
EXPAND ACT U -> A=(6,6)
EXPAND ACT U -> A=(5,6)
BLOCKED ACT L -> (5,5) WALL
EXPAND ACT L -> A=(6,5)
BLOCKED ACT U -> (5,5) WALL
EXPAND ACT L -> A=(6,4)
EXPAND ACT U -> A=(5,4)
BLOCKED ACT R -> (5,5) WALL
EXPAND ACT L -> A=(5,3)
EXPAND ACT L -> A=(5,2)
GOAL_REACHED
PLAN_LEN 6
PLAN_STEP 1 U
PLAN_STEP 2 L
PLAN_STEP 3 L
PLAN_STEP 4 U
PLAN_STEP 5 L
PLAN_STEP 6 L
SEARCH_TRACE_END
\end{verbatim}

\noindent\textbf{Explicit format}
\begin{verbatim}
INSTANCE_BEGIN
GRID_SIZE 10
MAP
# # # # # # # # # #
# . . # # . # . . #
# . # . # . . . . #
# # # . . # . . . #
# # . . . . . # . #
# . G . . # . . . #
# . . . . . . . # #
# . # . . . S . # #
# . . . . . . # . #
# # # # # # # # # #
START (7,6)
GOAL (5,2)
INSTANCE_END
SEARCH_TRACE_BEGIN
EXPAND sid=0 A=(7,6)
EXPAND sid=0 ACT U -> sid=1 A=(6,6)
EXPAND sid=1 ACT U -> sid=2 A=(5,6)
BLOCKED ACT L -> (5,5) WALL
EXPAND sid=1 ACT L -> sid=3 A=(6,5)
BLOCKED ACT U -> (5,5) WALL
EXPAND sid=3 ACT L -> sid=4 A=(6,4)
EXPAND sid=4 ACT U -> sid=5 A=(5,4)
BLOCKED ACT R -> (5,5) WALL
EXPAND sid=5 ACT L -> sid=6 A=(5,3)
EXPAND sid=6 ACT L -> sid=7 A=(5,2)
GOAL_REACHED
PLAN_LEN 6
PLAN_STEP 1 U
PLAN_STEP 2 L
PLAN_STEP 3 L
PLAN_STEP 4 U
PLAN_STEP 5 L
PLAN_STEP 6 L
SEARCH_TRACE_END
\end{verbatim}

\subsection{Sokoban}

The map is rendered with \texttt{\#} walls, \texttt{.} floor, \texttt{G} goal, \texttt{B} box, and \texttt{P} player. The two examples below describe the same instance and the same underlying search (plan length 5, 13 expansions).

\noindent\textbf{Implicit format}
\begin{verbatim}
INSTANCE_BEGIN
SIZE 10 10
BOARD
# # # # # # # # # #
# . . . . . . . . #
# . . . . . B G . #
# . . . B P . . . #
# . # # G . . # # #
# # # # # # # # # #
# # # # # # # # # #
# # # # # # # # # #
# # # # # # # # # #
# # # # # # # # # #
GOALS {(2,7),(4,4)}
INIT P=(3,5) B={(2,6),(3,4)}
INSTANCE_END
SEARCH_TRACE_BEGIN
EXPAND P=(3,5) B={(2,6),(3,4)}
EXPAND ACT U -> P=(2,5) B={(2,6),(3,4)}
EXPAND ACT R -> P=(2,6) B={(2,7),(3,4)}
EXPAND ACT U -> P=(1,6) B={(2,7),(3,4)}
EXPAND ACT D -> P=(3,6) B={(2,7),(3,4)}
EXPAND ACT L -> P=(2,5) B={(2,7),(3,4)}
EXPAND ACT L -> P=(1,5) B={(2,7),(3,4)}
EXPAND ACT R -> P=(1,7) B={(2,7),(3,4)}
EXPAND ACT D -> P=(4,6) B={(2,7),(3,4)}
EXPAND ACT L -> P=(3,5) B={(2,7),(3,4)}
EXPAND ACT R -> P=(3,7) B={(2,7),(3,4)}
EXPAND ACT L -> P=(2,4) B={(2,7),(3,4)}
EXPAND ACT D -> P=(3,4) B={(2,7),(4,4)}
GOAL_REACHED
PLAN_LEN 5
PLAN_STEP 1 U
PLAN_STEP 2 R
PLAN_STEP 3 L
PLAN_STEP 4 L
PLAN_STEP 5 D
SEARCH_TRACE_END
\end{verbatim}

\noindent\textbf{Explicit format}
\begin{verbatim}
INSTANCE_BEGIN
SIZE 10 10
BOARD
# # # # # # # # # #
# . . . . . . . . #
# . . . . . B G . #
# . . . B P . . . #
# . # # G . . # # #
# # # # # # # # # #
# # # # # # # # # #
# # # # # # # # # #
# # # # # # # # # #
# # # # # # # # # #
GOALS {(2,7),(4,4)}
INIT P=(3,5) B={(2,6),(3,4)}
INSTANCE_END
SEARCH_TRACE_BEGIN
EXPAND sid=0 P=(3,5) B={(2,6),(3,4)}
EXPAND sid=0 ACT U -> sid=1 P=(2,5) B={(2,6),(3,4)}
EXPAND sid=1 ACT R -> sid=2 P=(2,6) B={(2,7),(3,4)}
EXPAND sid=2 ACT U -> sid=3 P=(1,6) B={(2,7),(3,4)}
EXPAND sid=2 ACT D -> sid=4 P=(3,6) B={(2,7),(3,4)}
EXPAND sid=2 ACT L -> sid=5 P=(2,5) B={(2,7),(3,4)}
EXPAND sid=3 ACT L -> sid=6 P=(1,5) B={(2,7),(3,4)}
EXPAND sid=3 ACT R -> sid=7 P=(1,7) B={(2,7),(3,4)}
EXPAND sid=4 ACT D -> sid=8 P=(4,6) B={(2,7),(3,4)}
EXPAND sid=4 ACT L -> sid=9 P=(3,5) B={(2,7),(3,4)}
EXPAND sid=4 ACT R -> sid=10 P=(3,7) B={(2,7),(3,4)}
EXPAND sid=5 ACT L -> sid=11 P=(2,4) B={(2,7),(3,4)}
EXPAND sid=11 ACT D -> sid=12 P=(3,4) B={(2,7),(4,4)}
GOAL_REACHED
PLAN_LEN 5
PLAN_STEP 1 U
PLAN_STEP 2 R
PLAN_STEP 3 L
PLAN_STEP 4 L
PLAN_STEP 5 D
SEARCH_TRACE_END
\end{verbatim}

\section{Alternative Reward Designs}\label{app:reward-alternatives}

Here we discuss two alternative reward function designs we considered but found to have practical drawbacks that motivated the bounded discounted form we use in the main paper.

\paragraph{Constant per-step penalty with a cap.}
The first alternative subtracts a constant per-step penalty from the correctness reward, capped at zero so the reward stays non-negative:
\begin{equation}
\label{eq:reward-cap}
R_{\text{cap}}(\tau) = \mathbf{1}[\mathrm{valid}(\tau) \land \mathrm{correct}(\tau)]\,\max\!\left(1 - \alpha\,N_{\mathrm{exp}}(\tau),\, 0\right),
\end{equation}
where $\alpha > 0$ is the per-step penalty. This reward is sensitive to the choice of $\alpha$: the issue is that once $N_{\mathrm{exp}}(\tau) > 1/\alpha$ the reward saturates at zero and no longer discriminates between traces of different lengths. This also leads to some correct traces having the same reward as incorrect traces. While tuning $\alpha$ can mitigate this issue, the training was still unstable and tend to collapse after a few steps.

\paragraph{Geometric decay applied to the correctness reward.}
The second alternative multiplies the correctness reward by a geometric decay factor:
\begin{equation}
\label{eq:reward-decay}
R_{\text{decay}}(\tau) = \mathbf{1}[\mathrm{valid}(\tau) \land \mathrm{correct}(\tau)]\,\gamma^{N_{\mathrm{exp}}(\tau)},
\end{equation}
which avoids the saturation issue but still suffer from unstable training in our experiments. One explanation is the large range of scores assigned across correct traces: a difference of a few extra expansions changes the reward by an order of magnitude, so correct-but-long traces are penalized very heavily and contribute high-variance advantages during GRPO updates.

\newpage

\section{Visualization of Reasoning Environments}

\begin{figure}[H]
\centering

\begin{subfigure}[b]{\textwidth}
\centering
\begin{tabular}{@{}c@{\hspace{0.04\textwidth}}c@{}}
\exampleimage{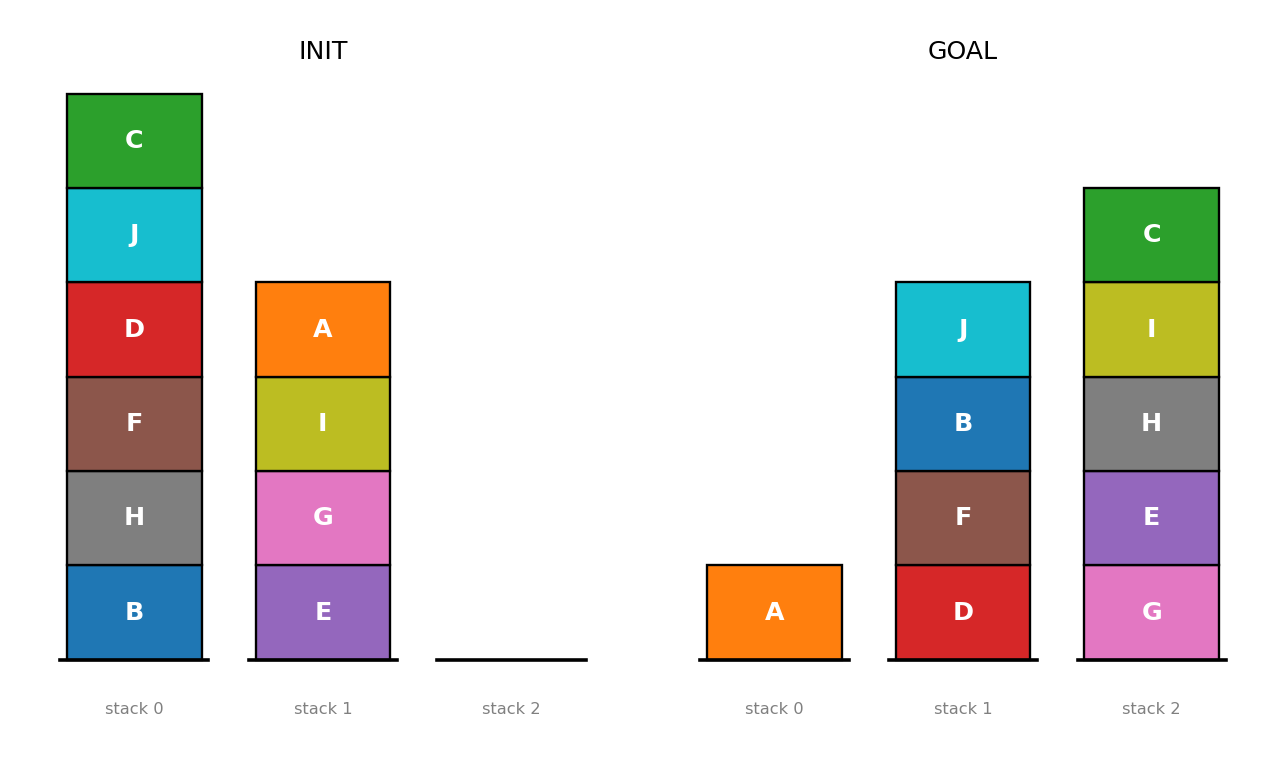}
&
\exampletext{
{\sffamily\bfseries Prompt}\\
INSTANCE\_BEGIN\\
BLOCKS A,B,C,D,E,F,G,H,I,J\\
INIT S\{ 0:B<H<F<D<J<C ; 1:E<G<I<A \}\\
GOAL S\{ 0:A ; 1:D<F<B<J ; 2:G<E<H<I<C \}\\
INSTANCE\_END\\
SEARCH\_TRACE\_BEGIN\\[0.7em]
{\sffamily\bfseries State representation in search}\\
"S\{ 0:B<H<F<D<J<C ; 1:E<G<I<A \}"
} \\
\end{tabular}
\caption{Blocks World}
\end{subfigure}

\vspace{0.4em}

\begin{subfigure}[b]{\textwidth}
\centering
\begin{tabular}{@{}c@{\hspace{0.04\textwidth}}c@{}}
\exampleimage{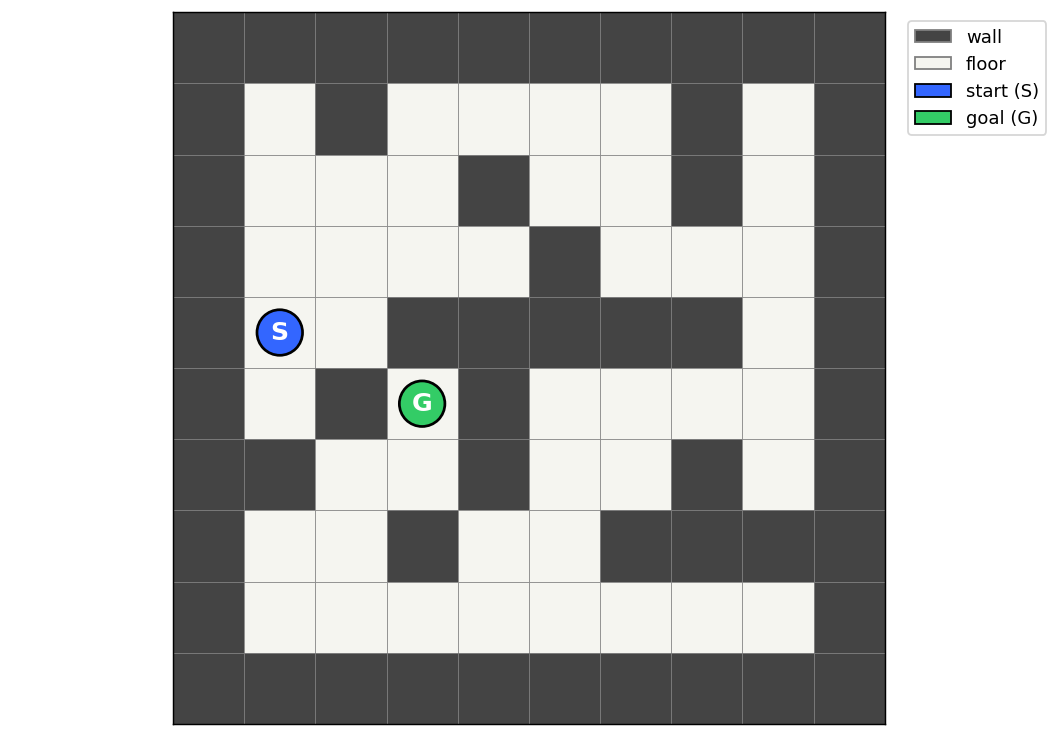}
&
\exampletext{
{\sffamily\bfseries Prompt}\\
INSTANCE\_BEGIN\\
GRID\_SIZE 10\\
MAP\\
\#~\#~\#~\#~\#~\#~\#~\#~\#~\#\\
\#~.~\#~.~.~.~.~\#~.~\#\\
\#~.~.~.~\#~.~.~\#~.~\#\\
\#~.~.~.~.~\#~.~.~.~\#\\
\#~S~.~\#~\#~\#~\#~\#~.~\#\\
\#~.~\#~G~\#~.~.~.~.~\#\\
\#~\#~.~.~\#~.~.~\#~.~\#\\
\#~.~.~\#~.~.~\#~\#~\#~\#\\
\#~.~.~.~.~.~.~.~.~\#\\
\#~\#~\#~\#~\#~\#~\#~\#~\#~\#\\
START (4,1)\\
GOAL (5,3)\\
INSTANCE\_END\\
SEARCH\_TRACE\_BEGIN\\[0.7em]
{\sffamily\bfseries State representation in search}\\
"A=(4,1)"
} \\
\end{tabular}
\caption{Navigation}
\end{subfigure}

\vspace{0.4em}

\begin{subfigure}[b]{\textwidth}
\centering
\begin{tabular}{@{}c@{\hspace{0.04\textwidth}}c@{}}
\exampleimage{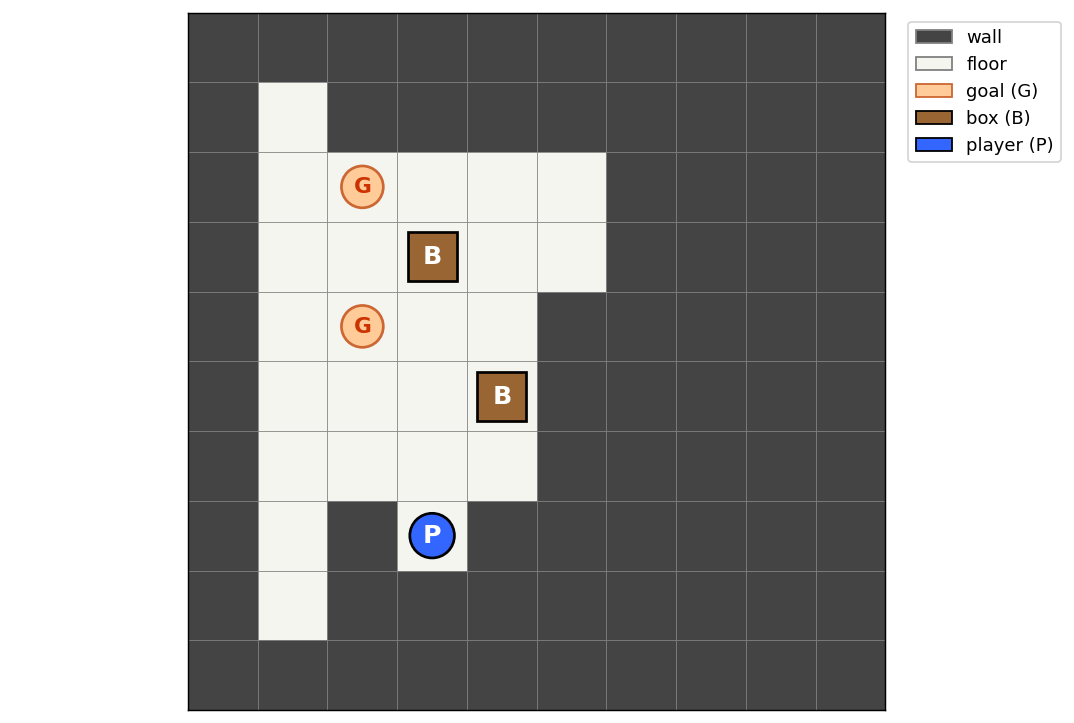}
&
\exampletext{
{\sffamily\bfseries Prompt}\\
INSTANCE\_BEGIN\\
SIZE 10 10\\
BOARD\\
+~+~+~+~+~+~+~+~+~+\\
+~.~+~+~+~+~+~+~+~+\\
+~.~O~.~.~.~+~+~+~+\\
+~.~.~.~.~.~+~+~+~+\\
+~.~O~.~.~+~+~+~+~+\\
+~.~.~.~.~+~+~+~+~+\\
+~.~.~.~.~+~+~+~+~+\\
+~.~+~.~+~+~+~+~+~+\\
+~.~+~+~+~+~+~+~+~+\\
+~+~+~+~+~+~+~+~+~+\\
GOALS \{(2,2),(4,2)\}\\
INIT P=(7,3) B=\{(3,3),(5,4)\}\\
INSTANCE\_END\\
SEARCH\_TRACE\_BEGIN\\[0.7em]
{\sffamily\bfseries State representation in search}\\
"P=(7,3) B=\{(3,3),(5,4)\}"
} \\
\end{tabular}
\caption{Sokoban}
\end{subfigure}

\caption{
Examples of the three domains used in our experiments. Each subfigure shows a visualization of one problem instance on the left and the corresponding textual serialization on the right.
}
\label{fig:domain-examples}
\end{figure}



\end{document}